\documentclass[runningheads]{llncs}

\usepackage{eccv}

\usepackage{eccvabbrv}

\usepackage{graphicx}
\usepackage{booktabs}

\usepackage[accsupp]{axessibility}  

\usepackage{hyperref}

\usepackage{orcidlink}

\begin{document}
\footnotetext[2]{$^\ast$ Corresponding authors.}
\title{InfMAE: A Foundation Model in The \\ Infrared Modality} 

\titlerunning{InfMAE: A Foundation Model in The Infrared Modality}

\author{Fangcen Liu\inst{1} \and
Chenqiang Gao\inst{2*} \and
Yaming Zhang\inst{1} \and
Junjie Guo\inst{1} \and \\
Jinghao Wang\inst{2} \and
Deyu Meng\inst{3}}

\authorrunning{F. Liu~ et al.}

\institute{School of Communications and Information Engineering, Chongqing University of Posts and Telecommunications, Chongqing 400065, China. \\
\email{\{liufc67,YummyZhang1989, gjj893866738\}@gmail.com} \and
School of Intelligent Systems Engineering, Sun Yat-sen University, Shenzhen, Guangdong 518107, China. \email{\{gaochq6,wangjh298\}@mail.sysu.edu.cn} \and
School of Mathematics and Statistics, Xi’an Jiaotong University, Xi’an, Shanxi, 710049, China, and School of Computer and Information Technology, Shanxi University, Taiyuan, Shanxi, China. \email{dymeng@mail.xjtu.edu.cn}
}

\maketitle

\begin{abstract}
In recent years, foundation models have swept the computer vision field, facilitating the advancement of various tasks within different modalities. 
However, effectively designing an infrared foundation model remains an open question.  
In this paper, we introduce InfMAE, a foundation model tailored specifically for the infrared modality. 
Initially, we present Inf30, an infrared dataset developed to mitigate the scarcity of large-scale data for self-supervised learning within the infrared vision community. 
Moreover, considering the intrinsic characteristics of infrared images, we design an information-aware masking strategy. 
It allows for a greater emphasis on the regions with richer information in infrared images during the self-supervised learning process, which is conducive to learning strong representations. 
Additionally, to enhance generalization capabilities in downstream tasks, we employ a multi-scale encoder for latent representation learning.
Finally, we develop an infrared decoder to reconstruct images.
Extensive experiments show that our proposed method InfMAE outperforms other supervised and self-supervised learning methods in three key downstream tasks: infrared image semantic segmentation, object detection, and small target detection.
\end{abstract}

\section{Introduction}
\label{sec:intro}
Infrared imaging plays a crucial role and finds extensive applications in low-light even dark and smoky conditions \cite{10088423, li2022dense, chen2021infrared} since they utilize the thermal radiation of objects for imaging, which can effectively alleviate adverse effects caused by challenging environmental conditions.
Currently, computer vision tasks in the infrared vision community, such as object detection \cite{liu2023infrared, 10288394}, and semantic segmentation \cite{li2023near,tian2023vu}, are explored by task-specific designed models.
While these models can yield superior performance on individual tasks, they are often challenged by their limited generalizability.

\begin{figure}
    \centering
    \includegraphics[width=1\textwidth]{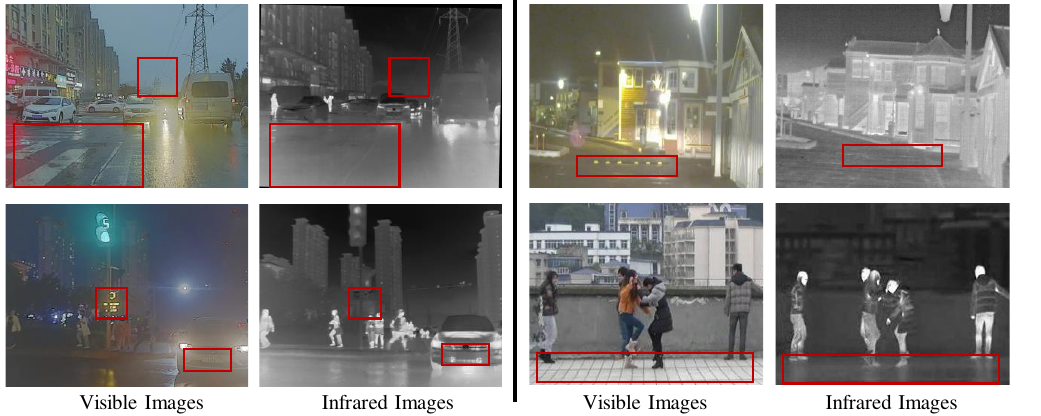}
    \caption{Visual comparison of infrared and visible images.
    Compared to visible images, infrared images display diminished informational content due to their inherent lack of rich texture and color details.
    For example, objects such as zebra crossings, telephone poles, and road signs in the red box are often submerged in their surroundings due to their similar temperature to the surrounding environment.
    }
    \label{fig:motivation}
\end{figure}

The foundation model refers to a large, general deep learning model that learns extensive knowledge through pre-training on a large-scale dataset and can be adapted to various downstream tasks by fine-tuning or prompting \cite{awais2023foundational, DBLP:journals/corr/abs-2108-07258, clip}.
The Transformer-based model like ViT \cite{vit} is one of the representative vision foundation models, which are trained on a large-scale dataset and outperforms the ResNet-based method \cite{kolesnikov2020big} on the downstream tasks.
To better learn the powerful representation of a foundation model from a large number of unlabeled data, self-supervised learning is the primary choice \cite{10234504}.
The self-supervised learning frameworks, such as BEiT \cite{beit}, DINO \cite{dino}, and Mask Autoencoders (MAE) \cite{mae}, achieve impressive performance across various downstream vision tasks \cite{vitdet, scheibenreif2023masked}.
Notably, the MAE \cite{mae} has verified superior representation learning ability and scalability.
This model is pre-trained on the ImageNet-1K dataset \cite{Imagenet}.
Then, the pre-trained encoder of the MAE \cite{mae} serves as a sufficient feature extraction backbone, achieving excellent performance on classification, detection, and segmentation tasks.
Following the self-supervised learning frameworks, pre-trained foundation models have emerged in various modalities, e.g., VideoMAE \cite{videomae, videomaev2} in the video modality, Scale-MAE \cite{Scale-mae} in the remote sensing image modality, and Point-M2AE \cite{Point-m2ae} in the point cloud modality.
\textit{However, a comparable foundation model has not yet emerged in the infrared modality.}

Intuitively, a feasible solution is to directly employ the pre-trained visible MAE on infrared downstream tasks.
However, as delineated in Section \ref{motivation}, our experimental findings suggest this is suboptimal, primarily due to significant disparities in imaging characteristics between infrared and visible modalities.
Fig. \ref{fig:motivation} illustrates these differences.
We can observe that infrared images exhibit less information compared to visible images, which lack rich texture and color details.
For instance, objects like zebra crossings and power lines in Fig. \ref{fig:motivation} lack color and texture information in infrared images, making them indistinguishable from the surrounding environment. 
These differences cause a noticeable gap between the infrared and visible modalities. 
The most straightforward strategy to mitigate this disparity is to train the MAE model on an extensive infrared dataset.
However, there appears to be a limited availability of a large-scale dataset in the infrared modality.
Consequently, it becomes important to build a large-scale infrared dataset to facilitate the training of an infrared foundation model.
In this paper, we begin by gathering a large number of infrared images.
Through image preprocessing like reducing redundancy, we finally obtain an infrared dataset named Inf30, which addresses the existing data scarcity within the infrared modality.

Due to the less information on infrared images, directly applying the vanilla MAE \cite{mae} architecture to the infrared modality is also a feasible but not the optimal choice, as the experiment shows in Sections \ref{m1}, \ref{m2} and \ref{m3}.
The vanilla MAE adopts the random mask strategy.
Directly applying this strategy to infrared images may cause the masking tokens to focus more on the information-poor part which is less conducive to representation learning.
Previous researches \cite{semmae, attmae} argued that masking tokens with rich semantic information could improve the representation learning ability of the self-supervised method, and introduced the semantic-guided masking strategy.
However, these methods do not account for the unique characteristics of the infrared modality and require the addition of new semantic-aware branches in the frameworks.

In this paper, we propose a simple yet effective foundation model in the infrared modality named InfMAE that can be extensively generalized to downstream tasks, such as infrared semantic segmentation, object detection, and small target detection.
To enhance the representation capabilities for infrared images, we design an information-aware masking strategy, which facilitates the selective masking of regions with richer information without designing an additional complex semantic awareness branch.
In addition, we employ a multi-scale strategy to enhance generalization capabilities in downstream tasks.
Finally, based on the characteristics of infrared images, we adopt a new infrared decoder module for image reconstruction.

We summarize the main contributions of this paper as follows:
\begin{itemize}
     \item[$\bullet$] We collect and preprocess a large number of infrared images, and finally release an infrared dataset Inf30 for self-supervised learning, which consists of 305241 infrared images.
     \item[$\bullet$] We propose the InfMAE, a simple yet effective foundation model in the infrared modality.
     To the best of our knowledge, this is the first attempt to build a foundation model within the infrared modality.
     \item[$\bullet$] Considering the characteristics of infrared imaging, we design an information-aware masking strategy and an infrared decoder. 
     These designs enhance the model's ability to learn strong representations.
     \item[$\bullet$] The experimental results show that the method proposed in this paper outperforms state-of-the-art methods, achieving the best performance in downstream tasks.
 \end{itemize}

\section{Related Work}
\label{sec:related work}
\subsection{The Foundation Model}
The foundation models are trained on a large-scale dataset in a self-supervised manner, making them adaptable for various downstream tasks \cite{DBLP:journals/corr/abs-2108-07258}.
The foundation model can be categorized into four categories based on different inputs: traditional models, textually prompted models, visually prompted models, and heterogeneous modalities-based models \cite{awais2023foundational}.
Traditional foundational models usually take only images as input, such as ResNet \cite{resnet} based on convolutional neural networks, as well as ViT \cite{vit} based on the Transformer architecture.
The textually prompted models like CLIP \cite{clip} and GPT-4 \cite{gpt4} took the textual prompt and the image as the inputs, and the textual prompt established correlations between text and images, contributing to achieving zero-shot transfer.
The visually prompted models \cite{CLIPSeg, SAM} could be prompted by various prompt types, e.g., text, bounding boxes, or a semantical mask to get the target segmentation.
The heterogeneous modalities-based models usually combine the various prompt types and the aligned multiple paired modalities, e.g., image-audio, image-video, etc., to learn meaningful representations.
The ImageBind \cite{Imagebind} is the representative work.
In this study, we attempt to explore the traditional foundation model in the infrared modality.

\subsection{The Masked Image Modeling and Representation Learning}
The Masked Image Modeling (MIM) \cite{mae, beit, mcmae, 10273635} is an emerging self-supervised learning approach, i.e., by randomly masking a part of the input image and then reconstructing it.
This learning method is simple yet effective in learning representations with high generality from a large-scale of data for downstream takes.
BEiT \cite{beit} and MAE \cite{mae} are pioneering works.
BEiT \cite{beit} used a bidirectional encoder representation to recover the original visual token based on the corrupted image patches.
MAE \cite{mae} pre-trained the ViT encoder by randomly masking the image tokens, feeding the visible tokens into the encoder, and then reconstructing the masked tokens through the decoder.
Based on MAE, pre-trained foundation models based on MIM learning have emerged in various modalities, e.g., VideoMAE \cite{videomae, videomaev2} in the video modality, Scale-MAE \cite{Scale-mae} in the remote sensing image modality, and Point-M2AE \cite{Point-m2ae} in the point cloud modality.
However, a MIM-based foundation model is not yet available in the infrared modality.

In addition to extending MIM methods to other modalities, methods also focus on designing mask strategies and how to reconstruct images.
Regarding the masking strategy, methods like MST \cite{Mst} and ATTMask \cite{attmae} utilized attention maps to guide the masking strategy. 
SemMAE \cite{semmae} used semantic information in images to select mask regions. 
UM-PVT \cite{umpvt} adopted a unified mask approach for masking learning. 
Regarding image reconstruction, some works \cite{beit, Peco} adopted tokens produced by VQ-VAE \cite{VQ-VAE} or its variants.

In this study, we explore a new masking strategy and decoder structure for infrared image reconstruction, according to the characteristics of infrared images.

\begin{figure}
    \centering
    \includegraphics[width=1\textwidth]{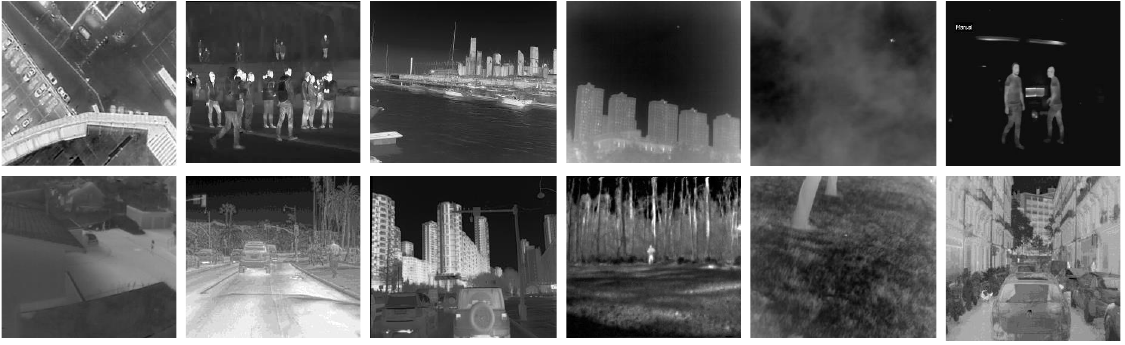}
    \caption{Some samples of the Inf30. 
    The environments in the collected dataset encompass skies, seascapes, forests, urban areas, suburban areas, lawns, and so on. 
    The objects include ships, vehicles, pedestrians, public facilities, residential buildings, and so on.
    }
    \label{fig:samples_of_dataset}
\end{figure}

\section{Pre-training Dataset Preparation: Inf30}
\label{sec:dataset}

\subsection{Collection and Preprocessing}
A large-scale dataset is essential for training the foundation model.
Hence, this study has constructed a dataset named Inf30.
The diversity of the images is important for learning the generalized representation. 
Hence, we extensively collect about 500K of infrared images from various websites \cite{he2018single, datasets, st2017mutual, datasets1, xu2019, datasets2, gao2016infar}.
However, the quality of these images is uneven.
For example, the images extracted from the same video exhibit significant similarity, leading to a reduction in image diversity.
Additionally, some infrared images obtained from the internet pose challenges due to their extremely low resolution.
Consequently, we initiate preprocessing all of the collected images to improve the quality of the infrared dataset.

As for the issue of high similarity in infrared images, we initially selected one image as the anchor.
Subsequently, we eliminated those images that have a closely resembling scene with the anchor image and retained images that exhibit substantial dissimilarity.
As for the issue of low resolution, we conducted data cleaning by removing images with both width and height of less than 20 pixels.

After collecting and preprocessing the infrared images, we finally obtained 305,241 images.
As illustrated in Fig. \ref{fig:samples_of_dataset}, the scene in our dataset encompasses skies, seascapes, forests, urban areas, suburban areas, lawns, and schools, while the objects include various types of ships, vehicles, pedestrians, public facilities, and residential buildings.
The minimum resolution of images is 40 $\times$ 23, and the maximum resolution of images is 6912 $\times$ 1024.

\subsection{Detail Information Analysis}
In this section, we analyze the detail richness of our Inf30 and the well-established ImageNet-1K \cite{Imagenet}. 
We employ the information entropy to measure the information of an image, which can reflect the detail richness.
The information entropy is defined as follows:

\begin{equation}
H(X) = - \sum_{i=1}^{n}P(x_i) \cdot log_2 (P(x_i)),
\end{equation}
where $n$ is the number of brightness levels in the image, $x_i$ represents each brightness level, and $P(x_i)$ is the probability of the brightness level in the image. 
We calculate the information entropy of images in the Inf30 and ImageNet-1K datasets.
The average information entropy of Inf30 is 6.44, while the average information entropy of ImageNet-1K is 7.19.
This reveals that the visible dataset exhibits richer information compared to the infrared image dataset. 

Recognizing the low information richness of infrared images, this study introduces an information-aware masking strategy to enhance the generalized representation learning ability of the foundation model.

\section{Method}
\label{sec:method}
\subsection{Overview}
\begin{figure*}
    \centering
    \includegraphics[width=1.0\textwidth]{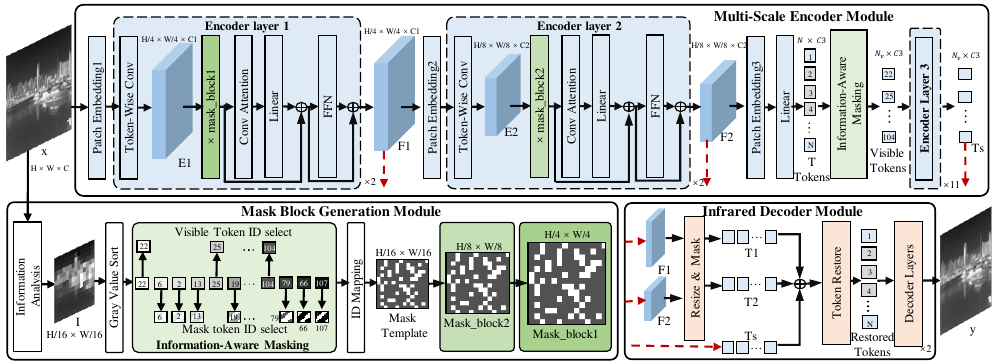}
    \caption{The framework of the proposed InfMAE. 
    It contains three modules: the mask block generation module, the multi-scale encoder module, and the infrared decoder module.
    The features connected by the red dashed lines in the multi-scale encoder module are fed into the infrared decoder module.
    }
    \label{fig:pipline}
\end{figure*}

As depicted in Fig. \ref{fig:pipline}, the proposed InfMAE architecture consists of three principal modules: 1) the mask block generation module, 2) the multi-scale encoder module, and 3) the infrared decoder module.
In the mask block generation module, we adopt an information-aware masking strategy on input tokens to get visible token IDs, mask\_block1, and mask\_block2, which are used in the multi-scale encoder module.
In the multi-scale encoder module, inspired by the MCMAE \cite{mcmae}, we combine convolution and self-attention mechanisms to conduct multi-scale representation learning for the visible tokens. 
In the infrared decoder module, we integrate the multi-scale representation with the learned visible tokens for image reconstruction.

\subsection{The Structure of the Proposed InfMAE}

\subsubsection{The Mask Block Generation Module}
Since infrared images lack color and texture details, employing a random masking strategy, as done by the vanilla MAE \cite{mae}, would result in excessive masking of information-poor regions.
This could undermine the network's capacity to develop robust representations from infrared images. 
Hence, this study proposes an information-aware masking strategy that can dynamically adjust mask tokens according to the image context.

Given an input image $x \in \mathbb{R}^{H \times W \times C}$, where $C$, $H$, and $W$ denote the number of channels, height, and width respectively.
$x$ is first input into the information analysis part to get the intensity-wise feature map. 
The information analysis part is a convolution with both the kernel size and stride set to 16.
Then we calculate the mean value of the intensity-wise feature map along the channel dimension and obtain the gray value map $I \in \mathbb{R}^{\frac{H}{16} \times\frac{W}{16}}$.
After that, the gray value map is flattened to $N$ tokens, $N = {\frac{H}{16} \times\frac{W}{16}}$.
Meanwhile, these tokens are arranged in descending order of their gray values, where a higher value means that the information is richer.
As depicted in the lower left corner of Fig. \ref{fig:pipline}, each box represents a token, and the number within the box is the sorted token ID.
After that, we perform the information-aware masking strategy on the sorted tokens, with a sampling stride $S$. 
Specifically, we sample the sorted tokens at an interval of $S$.
The sampled tokens are the visible tokens and the rest are masked tokens.
Meanwhile, we record the orders of the visible tokens' IDs and the masked tokens' IDs.
These IDs are used as a reference for generating mask blocks.

For mask block generation, we first obtain a binary value mask template through the ID mapping according to the sorted tokens IDs.
In the mask template in Fig. \ref{fig:pipline}, the white area is the visible token while the black area is the masked token.
Then we up-sample the mask template two and four times to get the mask\_block2 and mask\_block1 for the following multi-scale learning in the multi-scale encoder module.

\subsubsection{The Multi-scale Encoder Module}
\label{encoder}
Inspired by previous works like MCMAE \cite{mcmae} and ConvNeXt \cite{convnextv2}, we introduce a multi-scale encoder for visible token learning. 
This multi-scale encoder module mainly contains three encoder layers, which are two CNN-based encoders, named encoder layer 1 and encoder layer 2, and one Transorfmer-based encoder, named encoder layer 3.
Such three encoder layers can output feature maps with multi-scale spatial resolution, which are  $F_1 \in \mathbb{R}^{\frac{H}{4} \times\frac{W}{4} \times {C_1}}$, $F_2 \in \mathbb{R}^{\frac{H}{8} \times\frac{W}{8} \times {C_2}}$, and $T_s \in {N_v} \times {C_3}$, respectively. 
The $N_v$ is the number of the visible tokens.
Specifically, we first input the image $x$ into the patch embedding 1 and encoder layer 1 to obtain the feature map $F_1$.
It should be noted that after the patch embedding 1 and token-wise convolution, we obtain the feature map $E_1 \in \mathbb{R}^{\frac{H}{4} \times\frac{W}{4} \times {C_1}}$, and then we dot ${E_1}$ and mask\_block1 which are obtained in mask block generation module to mask the original feature.
Finally, we follow \cite{mcmae} and adopt the convolutional attention (Conv Attention) and FFN.
Same as encoder layer 1, the $F_1$ is inputted into patch embedding 2 and encoder layer 2 to obtain the feature maps $F_2 \in \mathbb{R}^{\frac{H}{8} \times\frac{W}{8} \times {C_2}}$.

After the patch embedding 3 and a linear operation, the $F_2$ is flattened into $N$ tokens $T \in {N} \times {C_3}$.
Then, we select the visible tokens from the $N$ tokens according to the visible token IDs obtained by the mask block generation module.
Finally, we use the encoder layer 3 whose structure is the same as the ViT \cite{vit} encoder layer, and obtain the learned visible tokens $T_s \in {N_v} \times {C_3}$.

\subsubsection{The Infrared Decoder Module}
In this study, we combine the multi-scale visible tokens and mask tokens and take them into the decoder for image reconstruction. 
Specifically, we normalize the multi-scale features ${F_1}$ and ${F_2}$ to the same size and then flatten them into the same shape with $T$.
After that, we obtain the visible tokens $T_1$ and $T_2$ according to the visible token IDs.
Then, we add all the values of visible tokens $T_1$, $T_2$, and $T_s$ and concat both masked tokens and visible tokens along the first dimension.
Meanwhile, we restore the order of these tokens based on the ID number in the mask block generation module.
We feed the restored tokens into the decoder for image reconstruction.
It is worth noting that infrared images lack many details and color information, so in this study, we set the infrared decoder depth to 2 for image reconstruction, whose structure is the same as the decoder of MAE \cite{mae}.
  
The loss function measures the Mean Square Error (MSE) between the masked patches of reconstructed image $y$ and the original image $x$.  
The MSE loss is defined as follows:

\begin{equation}
\mathcal{L}_{mse}=\frac{1}{M} \sum_{i=1}^{M} (y_i - x_i)^2,
\end{equation}
where the $M$ means the number of images.

\subsection{InfMAE for Downstream Tasks}
After the self-supervised pre-training, InfMAE learns multi-scale feature maps, which can be employed for existing infrared semantic segmentation, object detection, and small target detection methods.
To finetune InfMAE for these downstream tasks, we extract the $F_1$ and $F_2$ obtained by the encoder layer 1 and encoder layer 2.
Meanwhile, we reshape $T_s$ to the feature $F_3 \in \mathbb{R}^{\frac{H}{16} \times\frac{W}{16} \times {C_3}}$ and downsample the $F_3$ to $F_4 \in \mathbb{R}^{\frac{H}{32} \times\frac{W}{32} \times {C_4}}$ by a convolution layer.
It should be noted that when finetuning InfMAE for these downstream tasks, the shape of $T_s$ is ${N} \times {C_3}$. 
Finally, $F_1$, $F_2$, $F_3$, and $F_4$ are input into the downstream methods.

\section{Experiment}
\label{sec:experiment}

Initially, the proposed InfMAE is subjected to self-supervised pre-training using the released Inf30 dataset. 
Then, to assess the generalizability and robustness of InfMAE, it is applied to a series of downstream tasks. 
Finally, a series of ablation studies are conducted to confirm the efficacy and validity of the InfMAE framework.

\subsection{Pre-training Setup}
The framework of the proposed InfMAE is implemented using pytorch 1.12.0 and accelerated by CUDA 11.6 on 4 A100 GPUs.
The sampling stride $s$ of the information-aware masking strategy is set by default to 4.
The number of the encoder layer 1, encoder layer 2, and encoder layer 3 are set to 2, 2, and 11, respectively.
Meanwhile, the Transformer-based encoder layer 3 with 768 feature dimensions and 12 attention heads.
The infrared decoder is configured with two transformer layers, each having 512 feature dimensions and 12 attention heads. 
For training optimization, we adopt a 400-epoch cosine learning rate schedule, incorporating a 40-epoch warm-up phase. 
Optimization is achieved through the AdamW \cite{adam} optimizer, which is set to a base learning rate of $ 1.5 \times 10^{-4}$, a weight decay of 0.05, and a batch size of 256.
Additionally, random cropping is employed as a data augmentation strategy during pre-training.

\subsection{InfMAE on The Infrared Semantic Segmentation Task}
\subsubsection{Experimental Setup}
The MSRS \cite{MSRS} dataset is a semantic segmentation dataset of 1444 paired infrared and visible images.
The training set contains 1083 pairs of infrared and visible images and the testing set consists of 361 image pairs.
The images are categorized into eight major urban driving-related classes (Cars, People, Bicycles, Curves, Car Stops, Guardrails, Color Cones, and Bumps), providing a robust basis for urban scene analysis. 
In our experiments, we use the infrared component of the MSRS dataset, referred to as MARS-inf in the following description, to validate the performance of our method. 
We adopt UperNet \cite{upernet} and FCN \cite{fcn}, the hierarchical segmentation network headers, to compare the effectiveness of pre-trained InfMAE with other SOTA methods. 
We use a 240k iteration polynomial learning rate scheme with the first 1500 iterations warmed up. 
The AdamW \cite{adam} optimizer is used, with an initial learning rate of 10e-4, weight decay of 0.05, and a batch size of 2. 

\subsubsection{Results on The MSRS-inf}
\label{m1}
In our comparative analysis, the performance of the proposed method is evaluated against both supervised methods UperNet \cite{upernet}, DeeplabV3+ \cite{deeplabv3plus}, DNLNet \cite{dnl2020}, DDRNet \cite{DDRNet2022}, and self-supervised methods vanilla MAE \cite{mae}, MCMAE \cite{mcmae}, UM-MAE \cite{umpvt}.
We evaluate these methods using metrics such as the Intersection over Union (IoU) for each class, the mean Intersection over Union (mIoU) across all classes, and the mean accuracy (mAcc) for all classes.
The experimental results are shown in Table \ref{tab:semantic segmentation}. 
We first combine the encoder of the proposed InfMAE and the segmentation head of the Upernet\cite{upernet} and train the method from scratch in a supervised manner.
Compared with this method, leveraging the pre-trained weights of this encoder significantly enhances performance, as shown in Table \ref{tab:semantic segmentation}.
Meanwhile, compared with other supervised and self-supervised methods, the proposed method outperforms them a lot.

To evaluate the generalisability of the proposed InfMAE to other segmentation methods, we also use the FCN \cite{fcn} head for semantic segmentation.
As can be seen in Table \ref{tab:semantic segmentation}, the proposed method outperforms all the SOTA methods.

\begin{table*}[h]
\caption{Comparison with different semantic segmentation methods on the MSRS-inf \cite{MSRS} dataset, where Sup. (Scratch) indicates training our method from scratch.
        The bold and underline marks denote the best and suboptimal results, respectively.}
\label{tab:semantic segmentation}
\resizebox{\textwidth}{!}{%
\begin{tabular}{ccccccccccccc}
\hline
Methods        & Backbone & Model          & Car(\%)   & Person(\%) & Bike(\%)  & Curve(\%) & Car\_Stop(\%) & Guardrail(\%) & Color\_Cone(\%) & Bumb(\%)  & mIoU(\%)    & mAcc(\%)\\ \hline
UperNet \cite{upernet}       & Resnet50 & -                   & 82.6 & 67.2  & 64.2 & 47.3 & 55.1     & 56.6     & 53.9       & 65.9 & 65.6   & 74.7         \\
DeeplabV3+ \cite{deeplabv3plus}   & Resnet50 & -              & 84.2 & 69.5  & 63.7 & 45.1 & 54.7     & 68.8     & 44.5       & 58.8 & 65.2   & 73.8    \\
DNLNet \cite{dnl2020}          & Resnet101 & -                & 86.7 & 66.5  & 68.0 & 50.5 & 57.7     & 56.7     & 50.3       & 69.6 & 67.0   & 75.7   \\ 
DDRNet \cite{DDRNet2022}       & -       & -                  & 87.0 & 69.6  & 66.8 & 47.0 & 61.0     & 51.3     & 56.2       & 68.8 & 67.3   & 73.3 \\ \hline
Sup.(Scratch) & ViT-B    & FCN                                & 75.4 & 63.7  & 55.9 & 36.4 & 49.1     & 46.9     & 28.6       & 55.7 & 56.5   & 63.4 \\ 
Vanila MAE \cite{mae}   & ViT-B    & FCN                      & 83.2 & 67.3  & 62.6 & 44.5 & 58.1     & 57.7     & 45.0       & 60.5 & 64.1 & 71.6 \\ 
UM-MAE \cite{umpvt}       & VPT-S    & FCN                    & 80.3 & 67.3  & 61.4 & 41.1 & 54.2     & 59.1     & 38.7       & 61.9 & 62.4 & 70.1 \\ 
MCMAE \cite{mcmae}        & ViT-B    & FCN                    & \underline{86.5} & \underline{68.8}  & \underline{66.8} & \underline{55.0} & \underline{68.3}     & \underline{67.0}     & \underline{52.3}       & \underline{72.7} & \underline{70.6} & \underline{78.7} \\ 
InfMAE(Ours)         & ViT-B    & FCN         & \textbf{88.2}  & \textbf{70.1}  & \textbf{68.2} & \textbf{56.1} & \textbf{70.0}  & \textbf{67.5}  & \textbf{54.5} & \textbf{74.9} & \textbf{72.0} & \textbf{80.0} \\ \hline
Sup.(Scratch) & ViT-B    & UperNet                   & 74.0  & 50.7  & 41.2 & 49.7 & 35.6     & 40.5     & 44.0       & 55.9 & 61.1  & 61.1 \\
Vanilla MAE \cite{mae}   & ViT-B    & UperNet        & 86.6 & 72.3  & 66.0 & 56.2 & 69.7     & 70.2     & 54.9       & 68.1 & 71.4  & 78.2 \\
UM-MAE \cite{umpvt}         & VPT-S    & UperNet        & 87.2 & 72.1  & \underline{66.6} & 54.0 & 63.8     & 56.4     & 52.1       & 69.0 & 68.8  &76.3\\
MCMAE \cite{mcmae}        & ViT-B    & UperNet        & \underline{87.5}          & \textbf{73.2}  & {66.5}          & \underline{57.1}          & \underline{69.7}           & \underline{67.2}          & \underline{57.0}                & \underline{73.0}          & \underline{72.1}  & \underline{79.8} \\ 
InfMAE(Ours)        & ViT-B    & UperNet         & \textbf{89.3} & \underline{72.8}           & \textbf{68.8} & \textbf{59.1} & \textbf{72.1}  & \textbf{76.7} & \textbf{57.1}       & \textbf{74.2} & \textbf{74.3}  & \textbf{82.5} \\ \hline

\end{tabular}
}
\end{table*}

\subsection{InfMAE on The Infrared Object Detection Task}
\subsubsection{Experimental Setup}
The M3FD \cite{M3FD} dataset is an infrared and visible target detection dataset that contains 4200 images. 
The dataset labels six targets (People, Cars, Bus, Motor, Trucks, Lamps) during driving.
We use the infrared images from it for experimental performance validation.
In the following description, we call this used dataset M3FD-inf.
We follow the most common setting to use the detection heads in Mask R-CNN \cite{maskrcnn} and Cascade R-CNN \cite{cascade} and fine-tune the pre-trained InfMAE model.  
We follow most of the settings of the benchmark ViT \cite{vit}. 
We report the performance of the target detection model under 260k iteration cosine scheduling with a base learning rate of 10e-8 and a weight decay of 0.1.
The batch size is set to 2.

\begin{table*}[]
\caption{Performances of different object detection methods on the M3FD-inf \cite{M3FD} dataset, where Sup. (Scratch) indicates training our method from scratch.
The bold and underline marks denote the best and suboptimal results, respectively.}
\label{tab:Object Detection}
\resizebox{\textwidth}{!}{%
\begin{tabular}{ccccccccccc}
\cline{1-11}
Methods        & Backbone & Model    & People(\%) & Car(\%)    & Bus(\%)     & Motor(\%) & Lamp(\%) & Truck(\%) & mAP(\%)      & AP50(\%)  \\ \hline
DETR \cite{detr}         & ResNet101 &  -       & 78.5   & 88.1   & 87.8    & 75.8  & 65.9 & 78.1  & 46.7    & 79.0     \\ 
Sparse R-CNN \cite{Sparsercnn}     & ResNet50 & -        & 84.2   & 88.7   & 88.0    & 75.3  & 66.2 & 74.0  & 48.3    & 79.4         \\
DINO \cite{dino}          & Swin-L   & -        & 83.2   & 87.5   & 85.6    & 61.5  & 60.4 & 73.5  & 45.3    & 75.3         \\ 
YoloV8 \cite{yolov8}     & CSPDarkNet         & -        & 42.8   & 54.3   & 58.2    & 32.5  & 16.2 & 40.7  & 40.8    & 63.0      \\ \hline
Sup.(Scratch) & ViT-B    & Mask R-CNN &83.3       & 88.6   & 90.1    & 72.4      & 66.4     & 79.1    & 48.7  & 80.0                \\ 
Vanilla MAE \cite{mae}  & ViT-B    & Mask R-CNN & 84.8       & 89.6    & 93.1   & 76.0      & 76.5     & 80.5      & 51.4 & 83.4                 \\ 
MCMAE \cite{mcmae}        & ViT-B    & Mask R-CNN & \underline{87.8}   & \textbf{90.8}    & \underline{93.1}    & \textbf{87.7}     & \textbf{85.9}    & \textbf{84.9}     & \underline{55.7}   & \textbf{88.4}               \\ 
InfMAE(Ours)                     & ViT-B    & Mask R-CNN & \textbf{87.9}   & \underline{90.7}    & \textbf{94.6}    & \underline{87.1}      & \underline{85.2}     & \underline{83.2}      & \textbf{56.2}    & \underline{88.1}               \\  \hline
Sup.(Scratch) & ViT-B    & Cascade R-CNN  & 83.7   & 88.6    & 91.1    & 68.2      & 63.9     & 79.0       & 50.9   & 79.1                  \\
Vanilla MAE \cite{mae}  & ViT-B    & Cascade R-CNN  & 85.2   &  88.9   & \underline{94.3}   & 72.2      &  72.6   & 79.6     & 52.8   & 82.1                         \\
MCMAE  \cite{mcmae}       & ViT-B    & Cascade R-CNN  & \underline{85.6}       &  \textbf{90.4}         & {93.1}      & \underline{76.8}     & \underline{80.9}       &  \textbf{83.7}  & \underline{54.7} & \underline{85.1}                 \\ 
InfMAE(Ours)        & ViT-B    & Cascade R-CNN  & \textbf{85.7}       &  \textbf{90.4}         & \textbf{94.9}      & \textbf{81.9}     & \textbf{81.4}      & \underline{83.4}   &  \textbf{55.8}  & \textbf{86.3}                   \\ \hline

\end{tabular}
}
\end{table*}

\subsubsection{Results on The M3FD-inf}
\label{m2}
We compare the proposed method with the supervised methods DETR\cite{detr}, Sparse R-CNN \cite{Sparsercnn}, DINO \cite{dino}, YoloV8 \cite{yolov8} and with the self-supervised methods vanilla MAE \cite{mae}, MCMAE \cite{mcmae}.
We report the AP50 of each class, the mAP and AP50 of all classes' performance of all methods, and the experiment results are shown in Table \ref{tab:Object Detection}.
As shown in this table, when combining the encoder of the proposed InfMAE and the detection head of the Cascade R-CNN \cite{cascade} method, the proposed method outperforms all supervised methods.
As for the self-supervised methods, the proposed method outperforms most of the SOTA methods.
We can observe a similar experiment performance when using the head of Mask R-CNN method.

\subsection{InfMAE on The Infrared Small Target Detection Task}
\subsubsection{Experimental Setup}
The IRSTD-1k \cite{zhang2022isnet} is a well-established infrared small target detection dataset that contains 1000 infrared modal images with pixel-level annotations. 
We replace the encoder in \cite{liu2023infrared} with the multi-scale encoder module in InfMAE.
During the training phase, we adopt the AdamW \cite{adam} optimizer with an initial learning rate of 0.001, a weight decay of 0.05, and a batch size of 2, and run for 240k iterations.

\begin{table*}[]
\centering
\caption{Performances of different infrared small target detection methods on the IRSTD-1k dataset, where Sup. (Scratch) indicates training our method from scratch.
The bold and underline marks denote the best and suboptimal results, respectively.}
\label{tab:small target detction}
\begin{tabular}{ccccccc}
\cline{1-6}
Methods                                                   & Backbone   & $P_{d}$(\%) & AUC(\%) & $F1$(\%)  & IoU(\%) \\ \hline
LPNetGA \cite{lpnetga}                                                 &  -          &   77.1     &    60.6     &  50.8         &  34.1  \\
DNANet \cite{li2022dense}                                                  &   -         & 89.6       &  87.8       &  76.4         & 61.8   \\
ISTR \cite{liu2023infrared}                                              & ViT-B      &  85.8           &   86.1           &  69.4       & 53.1   \\ \hline
Sup.(Scratch)                                            & ViT-B      & 86.1       & 86.1        &  72.6            &  57.0       \\
Vanilla MAE  \cite{mae}                                            & ViT-B      &  86.8      &  86.8       &   72.8           &  57.2       \\
MCMAE \cite{mcmae}                                                   & ViT-B      & 74.2       &  \underline{90.8}       &  \underline{78.4}            & \underline{64.5}        \\
UM-VPT \cite{umpvt}                                                  & VPT-S      &  \textbf{99.6}      & 88.9        &  73.5            & 58.1        \\ 
InfMAE(Ours)                                                    & ViT-B      &  \underline{96.6}      & \textbf{91.2}        & \textbf{79.5}             & \textbf{66.0}        \\ \hline
\end{tabular}
\end{table*}

\subsubsection{Results on The IRSTD}
\label{m3}
We compare the proposed method with the supervised methods LPNetGA \cite{lpnetga}, DNANet \cite{li2022dense}, ISTR \cite{liu2023infrared}, and with the self-supervised methods Vanilla MAE \cite{mae}, MCMAE \cite{mcmae}, UM-PVT \cite{umpvt}.
According to the \cite{ipi,liu2023infrared}, we report four main performance metrics: positive detection $P_{d}$, AUC, pixel-level F1 score $F1$, and IoU in the table \ref{tab:small target detction}.
From this table, we can see that the proposed foundation model outperforms all the SOTA methods in metrics of AUC, $F1$, and IoU.

\subsection{Ablation Study of The InfMAE}

\subsubsection{Module Ablation}
In this section, we validate the effect of the information-aware masking strategy and multi-scale strategy.
We adopt the UperNet \cite{upernet} as the segmentation head and the Mask R-CNN \cite{maskrcnn} as the object detection head.
The results are shown in Table \ref{tab:Module}.
From this Table, we can see that combining the proposed information-aware masking strategy and the introduced multi-scale strategy can achieve better performance in downstream tasks.

\begin{table}[]
\centering
\caption{Ablation study on the influence of the masking strategy and multi-scale strategy. The IAM means the information-aware masking strategy.}
\label{tab:Module}
\begin{tabular}{cccccc}
\hline
IAM   & Multi-scale      & Seg\_mIoU(\%)  & Seg\_mAcc(\%)  & Det\_AP50(\%) & Det\_mAP(\%) \\ \hline
-           & -            &  71.4   & 78.2   & 81.5  & 49.1  \\
\checkmark  & -            &  72.0   & 79.4   & 78.9  & 50.1       \\
-           & \checkmark   &  72.1   & 79.8   & 86.3  & 55.0  \\
\checkmark  & \checkmark   &  \textbf{74.3}   & \textbf{82.5}   & \textbf{88.1}  & \textbf{56.2}             \\ \hline
\end{tabular}%
\end{table}

\subsubsection{Decoder Depth}
In this section, we analyze the effects of different decoder depths on the experimental performance of downstream tasks.
We set the decoder depth to 2, 4, 8, and 12 respectively, and we also adopt the UperNet \cite{upernet} and the Mask R-CNN \cite{maskrcnn} as the segmentation and object detection heads.
The experimental performance is shown in the table \ref{tab:Decoder}.
From the table, we can see that segmentation and detection gain the best result when the decoder depth is 2.
From the experimental results, we can find that the infrared image does not need a deep decoder because of its lack of texture, detail, and color information.
Meanwhile, we can observe from this table that the semantic performance is more sensitive to the decoder depth.
\begin{table}[]
\centering
\caption{The influence of increasing decoder depths on the infrared semantic segmentation (UperNet) task and object detection (Mask R-CNN) task.}
\label{tab:Decoder}
\begin{tabular}{ccccc}
\hline
Decoder Depth   & Seg\_mIoU(\%) & Seg\_mAcc(\%) & Det\_AP50(\%) & Det\_mAP(\%) \\ \hline
2       & \textbf{74.3}   & \textbf{82.5} & \textbf{88.1} & \textbf{56.2} \\
4       & 72.9   & 80.4 & 87.9 & 55.6 \\   
8       & 73.2   & 80.7 & 87.6 & 53.8 \\ 
12      & 74.0   & 81.9 & 87.2 & 54.1  \\ \hline   
\end{tabular}
\end{table}

\subsubsection{Masking Strides}
In this section, we analyze the experimental performance under different masking strides. 
For the mask strides, we select 2, 4, 7 and 14, and the experimental results are shown in the table \ref{tab:Mask}.
From the table, we can see that when the masking stride is set to 4, the semantic segmentation metrics mIoU and mAcc are the best, while the detection metric AP50 is suboptimal.
Considering that setting the sampling stride to 2 leads to an increase in the model's parameter count, we select a masking stride of 4. 

\begin{table}[]
\centering
\caption{The influence of different masking strides of the information-aware masking strategy on the infrared semantic segmentation (UperNet) task and object detection (Mask R-CNN) task.}
\label{tab:Mask}
\begin{tabular}{ccccc}
\hline
Strides  & Seg\_mIoU(\%) & Seg\_mAcc(\%) & Det\_AP50(\%) & Det\_mAP(\%) \\ \hline
2     & 74.2   & 82.0   & \textbf{89.3} & 56.1 \\
4     & \textbf{74.3}   & \textbf{82.5} & 88.1 & \textbf{56.2}     \\
7     & 71.8   & 79.9 & 88.0 & 54.9 \\ 
14    & 70.4    & -   & -  & 54.3 \\ \hline
\end{tabular}
\end{table}

\subsubsection{Pre-training Epochs}
For MAE, 1600 pre-training epochs can significantly improve the quality of the learned representations. 
To explore this effect of pre-training time, we train InfMAE with 200, 400, 800, and 1600 epochs and finetune the pre-trained models on downstream tasks.
The experimental results can be seen in Table \ref{tab:Epochs}.
From this table, we observe improved performances across most downstream tasks with an increase in pre-training epochs.

\begin{table}[]
\centering
\caption{The influence of increasing pre-training epochs on the infrared semantic segmentation (UperNet) task and object detection (Mask R-CNN) task.}
\label{tab:Epochs}
\begin{tabular}{ccccc}
\hline
Pre-train Epochs & Seg\_mIoU(\%) & Seg\_mAcc(\%) & Det\_AP50(\%) & Det\_mAP(\%) \\ \hline
200      & 71.3    & 78.9  & 87.4 & 55.4     \\
400      & 74.3    & 82.5  & 88.1 & 56.2  \\
800      & 74.9    & 83.1  & \textbf{88.3} & 55.9     \\
1600     & \textbf{74.9}    & \textbf{83.2}  & 88.2 & \textbf{56.0}     \\ \hline
\end{tabular}
\end{table}

\begin{table}[]
\centering
\caption{The influence of different pre-train data scales on the infrared semantic segmentation (UperNet) task and object detection (Mask R-CNN) task.
}
\label{tab:Data scale}
\begin{tabular}{ccccc}
\hline
Dataset Scale       & Seg\_mIoU(\%) & Seg\_mAcc(\%) & Det\_AP50(\%) & Det\_mAP(\%) \\ \hline
Inf10               & 69.1   & 77.6 & 86.9    & 55.0  \\
Inf15               & 73.0   & -    & -       & 55.6  \\
Inf30               & \textbf{74.3}   & \textbf{82.5} & \textbf{88.1} & \textbf{56.2}     \\
Inf50               & 73.0   & 80.5 & 87.3 & 55.8  \\ \hline   
\end{tabular}
\end{table}

\subsubsection{Pre-training Dataset Size Analysis}
In this section, we present the impact of varying pre-training dataset sizes on downstream task performance, as documented in Table \ref{tab:Data scale}.
The Inf10 and Inf15 contain 100K and 150K images which are sampled from Inf30 and the Inf50 contains 500K unprocessed images.
The table indicates a progressive enhancement in performance with the increase in dataset size from 100K to 300K while from 300K to 500K does not yield a proportional increase. 
This phenomenon is likely due to the increasing similarity of images within the larger dataset, which results in a limited representation learning ability of the foundational model. 
The lack of diversity in the dataset restricts the model’s generalization ability to various downstream tasks.
Thus, the findings highlight the significance of dataset quality and diversity in pre-training for optimal model performance.

\subsection{Pre-trained on Inf30 and ImageNet-1K}
\label{motivation}
In this section, we discuss the importance of the large-scale infrared dataset in developing an infrared foundation model.
To this end, the architecture of the vanilla MAE \cite{mae} is utilized for pre-training by 400 epochs on both the Inf30 dataset and the IN1K-30 dataset, the latter comprising 300k visible images randomly sampled from the ImageNet-1K \cite{Imagenet} for fair comparison.
To assess the generalization capabilities of the pre-trained models, we engage in infrared semantic segmentation using the MSRS-inf \cite{MSRS} dataset with UperNet \cite{upernet}, and infrared object detection on the M3FD-inf \cite{M3FD} dataset utilizing Mask R-CNN \cite{maskrcnn}. 
This experimental setting is consistent with the MAE \cite{mae}. 
The mIoUs and AP50 of vanilla MAE pre-trained with IN1K-30 are 66.7\% and 78.2\%, while pre-trained with Inf30 are 71.6\% and 83.4\%.
These results show that the MAE \cite{mae} pre-trains on Inf30 exhibit stronger generalization capabilities in infrared downstream tasks. 
This observation underscores that the distribution variances indeed contribute to performance disparities. 
Hence, the release of a large-scale infrared dataset is crucial for the development of the foundation model.

\section{Conclusion}
In this paper, we propose a foundation model in the infrared modality, named InfMAE.
We collect a dataset containing 305,241 infrared images for self-supervised learning. 
Meanwhile, to enhance the model's capacity to learn the generalized representation, an information-aware masking strategy is proposed, which can make the network pay more attention to the reconstruction of the information-rich part of the image.
Besides, based on the characteristics of infrared images, we design a new decoder structure. 
Finally, we validate the effectiveness of the proposed method InfMAE in infrared semantic segmentation, object detection, and small target detection tasks. 
The experimental results show that our InfMAE can obtain better experimental results than other supervised and self-supervised methods. 
In addition, we discuss some parameters in the InfMAE. 
We hope that the proposed foundation model InfMAE can promote the development of the infrared vision community.

\section*{Acknowledgements}
This work is supported in part by the National Key R\&D Program of China  (2022YFA1004100), and in part by the National Natural Science Foundation of China (No. 62176035, 62201111, 12226004, 62272375), the Science and Technology Research Program of Chongqing Municipal Education Commission under Grant (No.KJZD-K202100606).

%
%
\bibliographystyle{splncs04}
\bibliography{main}

\begin{thebibliography}{10}
\providecommand{\url}[1]{\texttt{#1}}
\providecommand{\urlprefix}{URL }
\providecommand{\doi}[1]{https://doi.org/#1}

\bibitem{awais2023foundational}
Awais, M., Naseer, M., Khan, S., Anwer, R.M., Cholakkal, H., Shah, M., Yang, M.H., Khan, F.S.: Foundational models defining a new era in vision: A survey and outlook (2023)

\bibitem{beit}
Bao, H., Dong, L., Piao, S., Wei, F.: Beit: Bert pre-training of image transformers. In: International Conference on Learning Representations (2021)

\bibitem{DBLP:journals/corr/abs-2108-07258}
Bommasani, R., Hudson, D.A., Adeli, E., Altman, R., Arora, S., von Arx, S., Bernstein, M.S., Bohg, J., Bosselut, A., Brunskill, E., et~al.: On the opportunities and risks of foundation models. arXiv preprint arXiv:2108.07258  (2021)

\bibitem{cascade}
Cai, Z., Vasconcelos, N.: Cascade r-cnn: Delving into high quality object detection. In: Proceedings of the IEEE conference on computer vision and pattern recognition. pp. 6154--6162 (2018)

\bibitem{detr}
Carion, N., Massa, F., Synnaeve, G., Usunier, N., Kirillov, A., Zagoruyko, S.: End-to-end object detection with transformers. In: Proceedings of the European conference on computer vision (ECCV). pp. 213--229. Springer (2020)

\bibitem{dino}
Caron, M., Touvron, H., Misra, I., J{\'e}gou, H., Mairal, J., Bojanowski, P., Joulin, A.: Emerging properties in self-supervised vision transformers. In: Proceedings of the IEEE/CVF international conference on computer vision. pp. 9650--9660 (2021)

\bibitem{lpnetga}
Chen, F., Gao, C., Liu, F., Zhao, Y., Zhou, Y., Meng, D., Zuo, W.: Local patch network with global attention for infrared small target detection. IEEE Transactions on Aerospace and Electronic Systems  \textbf{58}(5),  3979--3991 (2022)

\bibitem{deeplabv3plus}
Chen, L.C., Zhu, Y., Papandreou, G., Schroff, F., Adam, H.: Encoder-decoder with atrous separable convolution for semantic image segmentation. In: Proceedings of the European conference on computer vision (ECCV). pp. 801--818 (2018)

\bibitem{chen2021infrared}
Chen, X., Gao, C., Li, C., Yang, Y., Meng, D.: Infrared action detection in the dark via cross-stream attention mechanism. IEEE Transactions on Multimedia  \textbf{24},  288--300 (2021)

\bibitem{Imagenet}
Deng, J., Dong, W., Socher, R., Li, L.J., Li, K., Fei-Fei, L.: Imagenet: A large-scale hierarchical image database. In: 2009 IEEE conference on computer vision and pattern recognition. pp. 248--255. Ieee (2009)

\bibitem{Peco}
Dong, X., Bao, J., Zhang, T., Chen, D., Zhang, W., Yuan, L., Chen, D., Wen, F., Yu, N., Guo, B.: Peco: Perceptual codebook for bert pre-training of vision transformers. In: Proceedings of the AAAI Conference on Artificial Intelligence. vol.~37, pp. 552--560 (2023)

\bibitem{vit}
Dosovitskiy, A., Beyer, L., Kolesnikov, A., Weissenborn, D., Zhai, X., Unterthiner, T., Dehghani, M., Minderer, M., Heigold, G., Gelly, S., et~al.: An image is worth 16x16 words: Transformers for image recognition at scale. In: International Conference on Learning Representations (2020)

\bibitem{gao2016infar}
Gao, C., Du, Y., Liu, J., Lv, J., Yang, L., Meng, D., Hauptmann, A.G.: Infar dataset: Infrared action recognition at different times. Neurocomputing  \textbf{212},  36--47 (2016)

\bibitem{ipi}
Gao, C., Meng, D., Yang, Y., Wang, Y., Zhou, X., Hauptmann, A.G.: Infrared patch-image model for small target detection in a single image. IEEE transactions on image processing  \textbf{22}(12),  4996--5009 (2013)

\bibitem{mcmae}
Gao, P., Ma, T., Li, H., Lin, Z., Dai, J., Qiao, Y.: Mcmae: Masked convolution meets masked autoencoders. Advances in Neural Information Processing Systems  \textbf{35},  35632--35644 (2022)

\bibitem{Imagebind}
Girdhar, R., El{-}Nouby, A., Liu, Z., Singh, M., Alwala, K.V., Joulin, A., Misra, I.: Imagebind one embedding space to bind them all. In: {IEEE/CVF} Conference on Computer Vision and Pattern Recognition, {CVPR} 2023, Vancouver, BC, Canada, June 17-24, 2023. pp. 15180--15190. {IEEE} (2023)

\bibitem{mae}
He, K., Chen, X., Xie, S., Li, Y., Doll{\'a}r, P., Girshick, R.: Masked autoencoders are scalable vision learners. In: Proceedings of the IEEE/CVF conference on computer vision and pattern recognition. pp. 16000--16009 (2022)

\bibitem{maskrcnn}
He, K., Gkioxari, G., Doll{\'a}r, P., Girshick, R.: Mask r-cnn. In: Proceedings of the IEEE international conference on computer vision. pp. 2961--2969 (2017)

\bibitem{resnet}
He, K., Zhang, X., Ren, S., Sun, J.: Deep residual learning for image recognition. In: 2016 {IEEE} Conference on Computer Vision and Pattern Recognition, {CVPR} 2016, Las Vegas, NV, USA, June 27-30, 2016. pp. 770--778. {IEEE} Computer Society (2016)

\bibitem{he2018single}
He, Z., Cao, Y., Dong, Y., Yang, J., Cao, Y., Tisse, C.L.: Single-image-based nonuniformity correction of uncooled long-wave infrared detectors: A deep-learning approach. Applied optics  \textbf{57}(18),  D155--D164 (2018)

\bibitem{attmae}
Kakogeorgiou, I., Gidaris, S., Psomas, B., Avrithis, Y., Bursuc, A., Karantzalos, K., Komodakis, N.: What to hide from your students: Attention-guided masked image modeling. In: European Conference on Computer Vision. pp. 300--318. Springer (2022)

\bibitem{SAM}
Kirillov, A., Mintun, E., Ravi, N., Mao, H., Rolland, C., Gustafson, L., Xiao, T., Whitehead, S., Berg, A.C., Lo, W., Doll{\'{a}}r, P., Girshick, R.B.: Segment anything. CoRR  \textbf{abs/2304.02643} (2023)

\bibitem{kolesnikov2020big}
Kolesnikov, A., Beyer, L., Zhai, X., Puigcerver, J., Yung, J., Gelly, S., Houlsby, N.: Big transfer (bit): General visual representation learning. In: Proceedings of the European conference on computer vision (ECCV). pp. 491--507. Springer (2020)

\bibitem{datasets1}
Lab, A.S.: Thermal infrared dataset. \url{https://projects.asl.ethz.ch/datasets/doku.php?id=ir:iricra2014}

\bibitem{li2022dense}
Li, B., Xiao, C., Wang, L., Wang, Y., Lin, Z., Li, M., An, W., Guo, Y.: Dense nested attention network for infrared small target detection. IEEE Transactions on Image Processing  \textbf{32},  1745--1758 (2022)

\bibitem{semmae}
Li, G., Zheng, H., Liu, D., Wang, C., Su, B., Zheng, C.: Semmae: Semantic-guided masking for learning masked autoencoders. Advances in Neural Information Processing Systems  \textbf{35},  14290--14302 (2022)

\bibitem{umpvt}
Li, X., Wang, W., Yang, L., Yang, J.: Uniform masking: Enabling mae pre-training for pyramid-based vision transformers with locality. arXiv preprint arXiv:2205.10063  (2022)

\bibitem{li2023near}
Li, Y., Liu, H., Tian, Z., Geng, W.: Near-infrared vascular image segmentation using improved level set method. Infrared Physics \& Technology  \textbf{131},  104678 (2023)

\bibitem{vitdet}
Li, Y., Xie, S., Chen, X., Dollar, P., He, K., Girshick, R.: Benchmarking detection transfer learning with vision transformers. arXiv preprint arXiv:2111.11429  (2021)

\bibitem{Mst}
Li, Z., Chen, Z., Yang, F., Li, W., Zhu, Y., Zhao, C., Deng, R., Wu, L., Zhao, R., Tang, M., et~al.: Mst: Masked self-supervised transformer for visual representation. Advances in Neural Information Processing Systems  \textbf{34},  13165--13176 (2021)

\bibitem{10234504}
Li, Z.Y., Gao, S., Cheng, M.M.: Sere: Exploring feature self-relation for self-supervised transformer. IEEE Transactions on Pattern Analysis and Machine Intelligence  \textbf{45}(12),  15619--15631 (2023)

\bibitem{10288394}
Lin, F., Ge, S., Bao, K., Yan, C., Zeng, D.: Learning shape-biased representations for infrared small target detection. IEEE Transactions on Multimedia pp. 1--12 (2023)

\bibitem{liu2023infrared}
Liu, F., Gao, C., Chen, F., Meng, D., Zuo, W., Gao, X.: Infrared small and dim target detection with transformer under complex backgrounds. IEEE Transactions on Image Processing  \textbf{32},  5921--5932 (2023)

\bibitem{M3FD}
Liu, J., Fan, X., Huang, Z., Wu, G., Liu, R., Zhong, W., Luo, Z.: Target-aware dual adversarial learning and a multi-scenario multi-modality benchmark to fuse infrared and visible for object detection. In: Proceedings of the IEEE/CVF Conference on Computer Vision and Pattern Recognition. pp. 5802--5811 (2022)

\bibitem{fcn}
Long, J., Shelhamer, E., Darrell, T.: Fully convolutional networks for semantic segmentation. In: Proceedings of the IEEE conference on computer vision and pattern recognition. pp. 3431--3440 (2015)

\bibitem{adam}
Loshchilov, I., Hutter, F.: Decoupled weight decay regularization. In: 7th International Conference on Learning Representations, {ICLR} 2019, New Orleans, LA, USA, May 6-9, 2019. OpenReview.net (2019), \url{https://openreview.net/forum?id=Bkg6RiCqY7}

\bibitem{CLIPSeg}
L{\"{u}}ddecke, T., Ecker, A.S.: Image segmentation using text and image prompts. In: {IEEE/CVF} Conference on Computer Vision and Pattern Recognition, {CVPR} 2022, New Orleans, LA, USA, June 18-24, 2022. pp. 7076--7086. {IEEE} (2022)

\bibitem{10273635}
Madan, N., Ristea, N.C., Ionescu, R.T., Nasrollahi, K., Khan, F.S., Moeslund, T.B., Shah, M.: Self-supervised masked convolutional transformer block for anomaly detection. IEEE Transactions on Pattern Analysis and Machine Intelligence  \textbf{46}(1),  525--542 (2024)

\bibitem{gpt4}
OpenAI: Gpt-4 technical report. PREPRINT  (2023)

\bibitem{datasets}
OTCBVS: Otcbvs benchmark dataset collection. \url{https://vcipl-okstate.org/pbvs/bench/index.html}

\bibitem{DDRNet2022}
Pan, H., Hong, Y., Sun, W., Jia, Y.: Deep dual-resolution networks for real-time and accurate semantic segmentation of traffic scenes. IEEE Transactions on Intelligent Transportation Systems  \textbf{24}(3),  3448--3460 (2022)

\bibitem{clip}
Radford, A., Kim, J.W., Hallacy, C., Ramesh, A., Goh, G., Agarwal, S., Sastry, G., Askell, A., Mishkin, P., Clark, J., et~al.: Learning transferable visual models from natural language supervision. In: International conference on machine learning. pp. 8748--8763. PMLR (2021)

\bibitem{Scale-mae}
Reed, C.J., Gupta, R., Li, S., Brockman, S., Funk, C., Clipp, B., Keutzer, K., Candido, S., Uyttendaele, M., Darrell, T.: Scale-mae: A scale-aware masked autoencoder for multiscale geospatial representation learning. In: Proceedings of the IEEE/CVF International Conference on Computer Vision. pp. 4088--4099 (2023)

\bibitem{datasets2}
Reporter, T.M.: Infrared camera finds 6-year-old lost in deep woods. \url{https://www.youtube.com/watch?v=-FajSFRlkIo}

\bibitem{scheibenreif2023masked}
Scheibenreif, L., Mommert, M., Borth, D.: Masked vision transformers for hyperspectral image classification. In: Proceedings of the IEEE/CVF Conference on Computer Vision and Pattern Recognition. pp. 2165--2175 (2023)

\bibitem{st2017mutual}
St-Charles, P.L., Bilodeau, G.A., Bergevin, R.: Mutual foreground segmentation with multispectral stereo pairs. In: Proceedings of the IEEE International Conference on Computer Vision Workshops. pp. 375--384 (2017)

\bibitem{Sparsercnn}
Sun, P., Zhang, R., Jiang, Y., Kong, T., Xu, C., Zhan, W., Tomizuka, M., Li, L., Yuan, Z., Wang, C., et~al.: Sparse r-cnn: End-to-end object detection with learnable proposals. In: Proceedings of the IEEE/CVF conference on computer vision and pattern recognition. pp. 14454--14463 (2021)

\bibitem{MSRS}
Tang, L., Yuan, J., Zhang, H., Jiang, X., Ma, J.: Piafusion: A progressive infrared and visible image fusion network based on illumination aware. Information Fusion  \textbf{83-84},  79--92 (2022)

\bibitem{tian2023vu}
Tian, Z., Liu, H., Li, Q.: Vu-net: A symmetric network-based method for near-infrared blood vessel image segmentation. In: International Conference on Man-Machine-Environment System Engineering. pp. 275--280. Springer (2023)

\bibitem{videomae}
Tong, Z., Song, Y., Wang, J., Wang, L.: Videomae: Masked autoencoders are data-efficient learners for self-supervised video pre-training. Advances in neural information processing systems  \textbf{35},  10078--10093 (2022)

\bibitem{yolov8}
ultralytics: ultralytics, 2023. \url{https://github.com/ultralytics/ultralytics}

\bibitem{VQ-VAE}
Van Den~Oord, A., Vinyals, O., et~al.: Neural discrete representation learning. Advances in neural information processing systems  \textbf{30} (2017)

\bibitem{videomaev2}
Wang, L., Huang, B., Zhao, Z., Tong, Z., He, Y., Wang, Y., Wang, Y., Qiao, Y.: Videomae v2: Scaling video masked autoencoders with dual masking. In: Proceedings of the IEEE/CVF Conference on Computer Vision and Pattern Recognition. pp. 14549--14560 (2023)

\bibitem{convnextv2}
Woo, S., Debnath, S., Hu, R., Chen, X., Liu, Z., Kweon, I.S., Xie, S.: Convnext v2: Co-designing and scaling convnets with masked autoencoders. In: Proceedings of the IEEE/CVF Conference on Computer Vision and Pattern Recognition. pp. 16133--16142 (2023)

\bibitem{upernet}
Xiao, T., Liu, Y., Zhou, B., Jiang, Y., Sun, J.: Unified perceptual parsing for scene understanding. In: Proceedings of the European conference on computer vision (ECCV). pp. 418--434 (2018)

\bibitem{xu2019}
Xu, Z., Zhuang, J., Liu, Q., Zhou, J., Peng, S.: Benchmarking a large-scale fir dataset for on-road pedestrian detection. Infrared Physics {\&} Technology  \textbf{96},  199--208 (2019)

\bibitem{dnl2020}
Yin, M., Yao, Z., Cao, Y., Li, X., Zhang, Z., Lin, S., Hu, H.: Disentangled non-local neural networks. In: Proceedings of the European conference on computer vision (ECCV). pp. 191--207. Springer (2020)

\bibitem{zhang2022isnet}
Zhang, M., Zhang, R., Yang, Y., Bai, H., Zhang, J., Guo, J.: Isnet: Shape matters for infrared small target detection. In: Proceedings of the IEEE/CVF Conference on Computer Vision and Pattern Recognition. pp. 877--886 (2022)

\bibitem{Point-m2ae}
Zhang, R., Guo, Z., Gao, P., Fang, R., Zhao, B., Wang, D., Qiao, Y., Li, H.: Point-m2ae: multi-scale masked autoencoders for hierarchical point cloud pre-training. Advances in neural information processing systems  \textbf{35},  27061--27074 (2022)

\bibitem{10088423}
Zhang, X., Demiris, Y.: Visible and infrared image fusion using deep learning. IEEE Transactions on Pattern Analysis and Machine Intelligence  \textbf{45}(8),  10535--10554 (2023)

\end{thebibliography}
\end{document}